\documentclass{article}

\usepackage{arxiv}

\usepackage[utf8]{inputenc} 
\usepackage[T1]{fontenc}    
\usepackage{hyperref}       
\usepackage{url}            
\usepackage{booktabs}       
\usepackage{amsfonts}       
\usepackage{nicefrac}       
\usepackage{microtype}      
\usepackage{lipsum}
\usepackage{fancyhdr}       
\usepackage{graphicx}       
\usepackage{pdflscape}
\usepackage{amsthm,amsmath,amssymb}
\usepackage{algorithm}
\usepackage[noend]{algpseudocode}
\usepackage{multirow}
\usepackage{commath}

\DeclareMathOperator*{\argmin}{arg\,min}

\graphicspath{{media/}}     

\title{Exact algorithms and heuristics for capacitated covering salesman problems
}

\author{
  Lucas Porto Maziero, Fábio Luiz Usberti, Celso Cavellucci \\
  Institute of Computing \\
  Universidade Estadual de Campinas \\
  Campinas\\
  \texttt{\{lucas.maziero, fusberti, celsocv\}@ic.unicamp.br} \\
}

\begin{document}
\maketitle

\begin{abstract}

This paper introduces the Capacitated Covering Salesman Problem (CCSP), approaching the notion of service by coverage in capacitated vehicle routing problems. In CCSP, locations where vehicles can transit are provided, some of which have customers with demands. The objective is to service customers through a fleet of vehicles based in a depot, minimizing the total distance traversed by the vehicles. CCSP is unique in the sense that customers, to be serviced, do not need to be visited by a vehicle. Instead, they can be serviced if they are within a coverage area of the vehicle. This assumption is motivated by applications in which some customers are unreachable (e.g., forbidden access to vehicles) or visiting every customer is impractical. In this work, optimization methodologies are proposed for the CCSP based on ILP (Integer Linear Programming) and BRKGA (Biased Random-Key Genetic Algorithm) metaheuristic. Computational experiments conducted on a benchmark of instances for the CCSP evaluate the performance of the methodologies with respect to primal bounds. Furthermore, our ILP formulation is extended in order to create a novel MILP (Mixed Integer Linear Programming) for the Multi-Depot Covering Tour Vehicle Routing Problem (MDCTVRP). Computational experiments show that the extended MILP formulation outperformed the previous state-of-the-art exact approach with respect to optimality gaps. In particular, optimal solutions were obtained for several previously unsolved instances.

\end{abstract}


\section{Introduction}

The Capacitated Vehicle Routing Problem (CVRP), initially proposed by Dantzig and Ramser \cite{dantzig1959truck}, is one of the most well-known problems in combinatorial optimization. The goal of CVRP is to service the demands of a set of customers through a set of vehicles located in a depot, minimizing the total distance travelled. Each vehicle must depart and return to the depot, and cannot service more than its capacity \cite{cordeau2007vehicle}.

\begin{figure}[h]
	\centering	\includegraphics[width=0.75\textwidth]{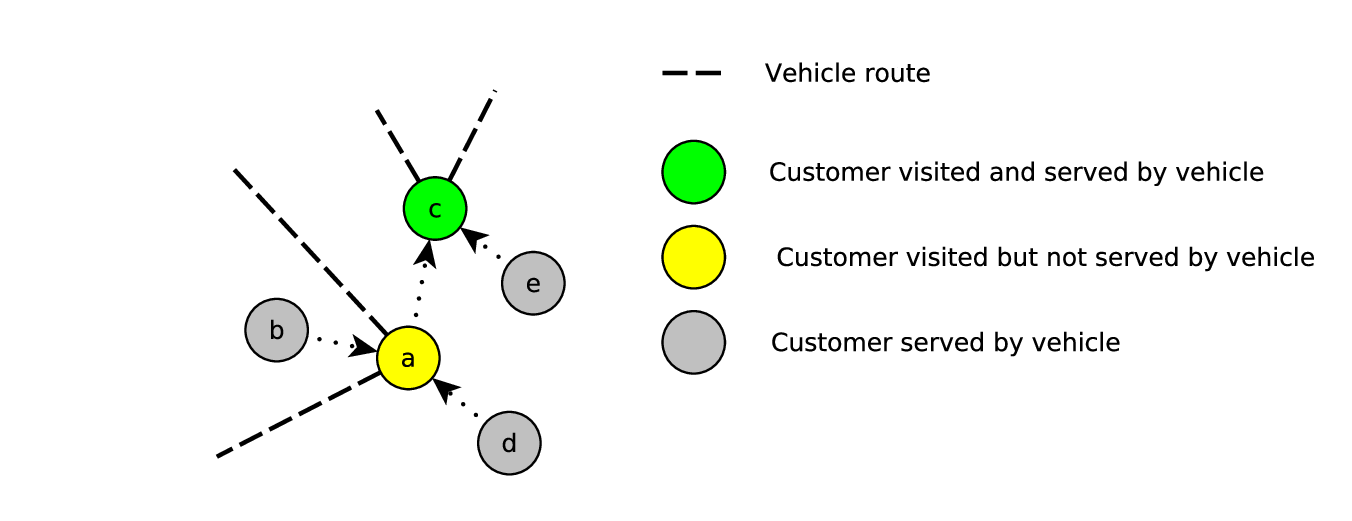} \\ [0.25cm]
	\caption{Example of covering ranges.}
	\label{fig:nocaoCobertura}
\end{figure}

The CVRP encompasses many variants with restrictions of time constraints, resources availability, and even customers accessibility, for example, regions of difficult means of entry to vehicles \cite{golden2008vehicle,allahyari2015hybrid}. The latter can be addressed by considering service by covering. A notion by which a customer can be serviced remotely as long as the customer is in the covering range of the vehicle. For example, in Figure~\ref{fig:nocaoCobertura} customers $b$ and $d$ are within the covering range of customer $a$. Also, customers $a$ and $e$ can be remotely serviced by vertex $c$, if there is enough remaining capacity in the corresponding vehicle.

The first problem using the concept of servicing by coverage is the Covering Salesman Problem (CSP), by Current and Schilling \cite{current1989coveringa2}, stated as follows. Given an undirected graph with cost attributed to the edges, the objective is to determine a minimum cost cycle such that every vertex out of the cycle is covered by at least one vertex in the cycle. 
The CSP generalizes the Travelling Salesman Problem (TSP) \cite{applegate2007a2} in the case where each vertex only covers itself, from which follows that CSP is NP-hard.

Generalizations of the CSP were investigated in literature. Golden et al.~\cite{golden2012generalizeda2} proposed a generalization of the CSP in which each vertex has a covering demand referring to the number of times it must be covered by the tour. Also, each vertex has a fixed cost that incurs from visiting it. The authors developed a heuristic with local search that explores exchange, removal, and insertion neighborhoods.

Gendreau et al. \cite{gendreau1997coveringa2} investigated the Covering Tour Problem (CTP), a problem where the vertices are categorized by those that can be visited $V$, must be visited $T \subseteq V$, and cannot be visited $W$. The goal of the CTP is to obtain a minimum cost Hamiltonian cycle over a set of vertices $S \subseteq V$ containing all vertices in $T$ and no vertices in $W$, and each vertex of $W$ is covered by at least one vertex in $S$. Exact and heuristic methodologies were proposed to solve the problem.

Hachicha et al. \cite{hachicha2000heuristics} introduced the Multi-Vehicle Covering Routing Problem ($m$-CTP). It generalizes the CTP in the sense that there are multiple vehicles, and each route cannot exceed predefined length and number of vertices. The $m$-CTP was used as the basis to formulate a problem of locating distribution centers for humanitarian aid in disaster areas \cite{naji2012covering}. Methodologies to solve the $m$-CTP include branch-and-cut \cite{ha2013exact}, column generation \cite{murakami2014column}, branch-and-price \cite{jozefowiez2014branch}, constructive heuristics \cite{hachicha2000heuristics}, evolutionary metaheuristic \cite{ha2013exact}, variable neighborhood descent\cite{kammoun2017integration}.

Allahyari et al. \cite{allahyari2015hybrid} proposed the the Multi-Depot Covering Tour Vehicle Routing Problem (MDCTVRP). The MDCTVRP is a combination of the Multi-Depot Vehicle Routing Problem (MDVRP) \cite{tillman1969multiple} and CSP. In the MDCTVRP, the demand of each customer can be served either by visiting the customer directly or by covering, i.e, the customer location is within a covering range of at least one visited customer. The authors developed two mixed integer programming formulations and a hybrid metaheuristic, combining Greedy Randomized Adaptive Search Procedure (GRASP), Iterated Local Search (ILS) and Simulated Annealing (SA).

It is worth noticing that the CSP does not have a multi-vehicle variant, as does the CTP. This work fills this gaps by proposing the Capacitated Covering Salesman Problem (CCSP), a NP-hard problem generalizing both the CVRP and the CSP. Vertices with non-negative demands must be covered by a set of capacitated vehicles, based at the depot. The goal is to find a minimum cost set of vehicle routes servicing all the demands. The CCSP represents a straightforward extension of the CSP where the service employed by the vehicles comes at a limited supply. At the same time, the CCSP generalizes the CVRP since covering provides an additional way to service each demand. It is worth pointing out the differences between $m$-CTP and CCSP:

\begin{itemize}
    \item CCSP considers demands on the vertices;
    \item $m$-CTP forces some vertices to be visited ($T \subseteq V$);
    \item $m$-CTP constrains the routes by their lengths and number of vertices, while in CCSP the vehicle is capacitated by the amount of serviced demand.
    \item CCSP is a natural generalization of the CSP and the $m$-CTP generalizes the CTP.
\end{itemize}

\paragraph{Our contributions} Two combinatorial optimization problems, the CSP and the VRP, are combined into a general framework to address routing problems with multiple vehicles and limited capacity in the context of service by covering. Mathematical formulations, using integer linear programming, are provided to represent these problems as a CCSP. The complexity of solving these problems optimally asks for heuristic methodologies to tackle large instances that arise from real applications. This work answers this demand by proposing a biased random-keys genetic algorithm to solve the CCSP, and a matheuristic to intensify the search. Furthermore, we extended our ILP to solve the MDCTVRP and conducted computational experiments on a benchmark of instances, comparing our formulation with the state-of-the-art exact methodology from literature. The proposed formulation outperformed the previous approach with respect to optimality gaps. Moreover, optimal solutions were proven for several previously unsolved instances.

The CCSP and MDCTVRP share common concepts of covering and vehicle capacity. Consequently, both can be modeled in a similar manner concerning serving remotely customers and demands served by the vehicle. Nevertheless, notable distinctions exist between these problems. In CCSP, a vehicle is not obligated to serve a customer during its visit, unlike in MDCTVRP, where the vehicle is required to serve a customer during each visit. Additionally, in CCSP there are customers with no specified demand, providing an opportunity for vehicles to use them to serve other customers remotely which have demand. In contrast, in MDCTVRP every customer is associated with a specific demand.

This paper is organized as follows. Section~\ref{sec:CCSP} formally defines the CCSP and MDCTVRP, presenting ILP formulations for the CCSP and a MILP formulation for the MDCTVRP. Section~\ref{sec:brkga} describes the BRKGA, and the intra-route and inter-route intensification procedures. In Section~\ref{sec:experimentsCCSP}, computational experiments are conducted on a representative set of instances, and results are analyzed and discussed. Section~\ref{sec:conclusionsCCSP} gives the concluding remarks.

\section{Mathematical Formulations} \label{sec:CCSP}

\subsection{Models for the CCSP}

Consider a complete undirected graph $G(V,E)$, where each vertex $v \in V$ has a demand $d_v$, each edge $e \in E$ has a metric cost $c_e$, and a depot vertex is denoted by $v_0$. Let $V_0 = V \setminus \{v_0\}$ and $V_d = \{v \in V : d_v > 0\}$. There are $M$ homogeneous vehicles with capacity $Q$ that must service all vertices with positive demand. 

For each vertex $v \in V$, $C(v)$ is the set of vertices that covers $v$ and $D(v)$ is the set of vertices that are covered by $v$. It is assumed that $v \in C(v)$ and $v \in D(v)$, $\forall v \in V$.

A route is a nonempty subset $R \subseteq E$ of edges for which the induced subgraph $G[R]$ is a simple cycle containing $v_0$. The goal of CCSP is to find $M$ routes of minimum cost with the following constraints:

\begin{itemize}
    \item each vertex is visited no more than once;
    \item each demand $d_v : v \in V_d$ is serviced by a route $R$, which implies in $v$ or some vertex in $C(v)$ being visited by $R$, and the demand $d_v$ being deducted from the capacity of the vehicle;
    \item the total demand serviced by any vehicle must not exceed its capacity $Q$.
\end{itemize}

\begin{figure}[h!]
	\centering
	\includegraphics[width=0.85\textwidth]{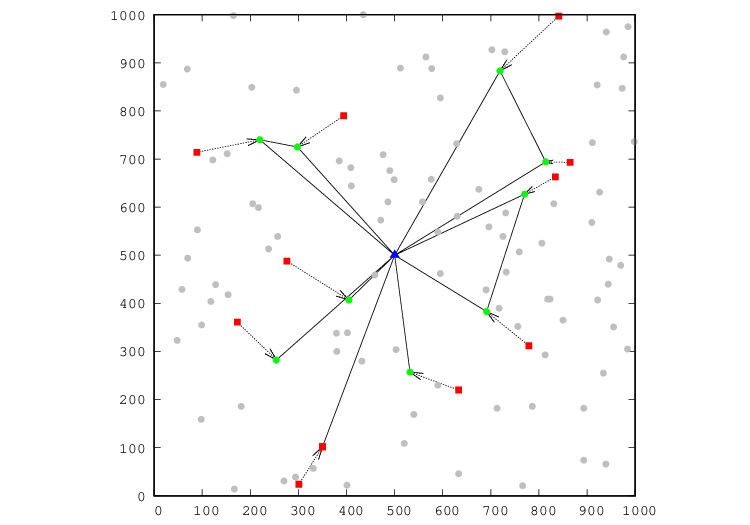} \\[0.125cm]
	\caption{Optimal solution for the CCSP instance $X$-$n115$-$w11$-$c7$.}
	\label{fig:solutionCCSP}
\end{figure}

Figure~\ref{fig:solutionCCSP} shows an optimal solution for a CCSP instance. Routes are depicted with black lines; the blue triangle is $v_0$; red squares are vertices with positive demand; green points are visited vertices with no demand; arrows show which route serviced each demand.

The following ILP formulation $CCSP_1$ is proposed for the CCSP. We denote $\delta(v)$ as the edge cut-set of vertex $v$, and $\delta(S)$ the edge cut-set of a subset $S \subseteq V$. The formulation includes the following decision variables: $x_e \in \mathbb{Z}^+$ gives the number of times edge $e \in E$ is traversed; $y_v \in \{0,1\}$ denotes if vertex $v \in V$ is visited (1) or not (0); $z_{uv} \in \{0,1\}$ shows if vertex $u \in V_d$ is serviced through vertex $v \in C(u)$ (1) or not (0); $K \in \mathbb{Z}^+$ is the number of vehicles.

\begin{align} \label{eq:modeloCCSP1}
	& (CCSP_1) \nonumber \\
	& MIN \quad \sum_{e \in E}{c_e x_e}, \\
	& \mbox{subject to} \nonumber \\
	& \sum_{e \in \delta(v_0)}{x_e} = 2K, \\
	& \sum_{e \in \delta(v)}{x_e} = 2y_v \qquad & \forall v \in V_0, \\
	& \sum_{v \in C(u)}{y_v} \geqslant 1 \qquad & \forall u \in V_d, \\
	& z_{uv} \leqslant y_v \qquad & \forall u \in V_d, \forall v \in C(u), \\
	& \sum_{v \in C(u)}{z_{uv}} = 1 \qquad & \forall u \in V_d, \\
	& \sum_{e \in \delta(S)}{x_e} \geqslant \frac{2}{Q} \sum_{u \in V_d}{\sum_{v \in (S \cap C(u))}{d_u z_{uv}}} \qquad & \forall S \subseteq V_0, \\	
	& x_{e} \in \{0,1\} \qquad & \forall e \notin \delta(v_0), \\
	& x_{e} \in \{0,1,2\} \qquad & \forall e \in \delta(v_0), \\
	& y_v \in \{0,1\} \qquad & \forall v \in V_0, \\
	& z_{uv} \in \{0,1\} \qquad & \forall u \in V_d, \forall v \in C(u), \\
	& K \in \mathbb{Z}^+.
\end{align}

The objective function ($1$) minimizes the total cost of the routes. Constraints ($2$) state that the vertex depot is visited by all $K$ routes. Constraints ($3$) ensure that the number of edges incident to a vertex $v \in V_0$ is $2$ if $v$ is visited or $0$ otherwise. Constraints ($4$) impose that each vertex in $V_d$ must be covered by at least one route. Constraints ($5$) state that if a vertex $u \in V_d$ is serviced by a vertex $v \in C(u)$, then $v$ is visited. Constraints ($6$) ensure that every vertex $u \in V_d$ is serviced by a vertex $v \in C(u)$. Constraints ($7$) impose both the connectivity and the vehicle capacity by forcing into the solution a sufficient number of edges to each subset of vertices.

A second formulation, denominated $CCSP_2$, can be derived by eliminating variables $y$ through variable substitution using constraints ($3$). 

\begin{align} \label{eq:modeloCCSP2}
	& (CCSP_2) \nonumber \\
	& MIN \quad \sum_{e \in E}{c_e x_e}, \nonumber \\
	& \mbox{subject to} \nonumber \\
	& \sum_{e \in \delta(v)}{x_e} \geqslant 2 z_{uv} \qquad & \forall v \in V_0, \forall u \in V_d, \\
	& \sum_{e \in \delta(v)}{x_e} \leqslant 2 \sum_{u \in V_d}{z_{uv}} \qquad & \forall v \in V_0, \\
	& \sum_{e \in \delta(v)}{x_e} \leqslant 2 \qquad & \forall v \in V_0, \\
	& (2), (6), (7), (8), (9), (11), (12). & \nonumber
\end{align}

Constraints ($13$), ($14$), and ($15$) impose the correct number of edges incident to a vertex $v \in V_0$ ($2$ if visited or $0$ otherwise).

Preliminary experiments have shown that, even though the $CCSP_2$ has fewer variables than the $CCSP_1$, the overall quality of the upper and lower bounds obtained by $CCSP_1$ is better than $CCSP_2$. Therefore, only the $CCSP_1$ formulation will be considered in the computational experiments.

\subsection{Models for a Multi-depot Variant}

The Multi-Depot Covering Tour Vehicle Routing Problem (MDCTVRP), proposed by Allahyari et al. \cite{allahyari2015hybrid}, is defined next. Given a directed graph $G = (N, A)$, with vertices $N = N_c \cup N_d$, and arcs $A$. Each customer $i \in N_c = \{1, 2, ..., n_c\}$ has a positive demand $d_i$. Set $N_d = \{1,...,n_d \}$ contains the depots. Each arc $(i, j) \in A$ has a positive traversing cost $c_{ij}$. Each customer has to be covered by a route. Set $C(v)$ represents the vertices that covers $v$. A cost $c'_{ij} > 0$ is attributed for servicing customer $i$ through $j$. A set of identical vehicles $P = \{1, 2, ..., p\}$ is available, and $Q$ is the vehicle capacity. Each depot $k \in N_d$ has a limited capacity $H$. Finally, to each depot is attributed a unique set $P_k = \{1, \ldots, p_k\}$ of vehicles. The objective of MDCTVRP is to find a minimum cost set of routes, such that all demands are covered, the vehicles and depots capacities are satisfied, and each vehicle starts and ends its route in the same depot.

Borrowing ideas from model $CCSP_1$ and from the flow-based formulation by Allahyari et al. \cite{allahyari2015hybrid}, we propose a new MILP formulation for the MDCTVRP. The formulation, henceforth denominated $MDCTVRP_m$, includes the following decision variables: $x_{ij}$ denotes if arc $(i, j) \in A$ is traversed ($1$) or not ($0$); $y_v$ represents if vertex $v \in N_c$ is visited ($1$) or not ($0$); $z_{uv}$ shows if vertex $u \in N_c$ is serviced by vertex $v \in N_c$ ($1$) or not ($0$); $f_{ij}$ gives the vehicle load while traversing arc $(i, j)$. We represent $SP$ as the set containing all simple paths connecting depots. Specifically, $SP_{(st)} \in SP$ denotes the set of all simple paths between depots $s$ and $t$.

\begin{align} \label{eq:mdctvrpModel}
	& (MDCTVRP_m) \nonumber \\
	& MIN \quad \sum_{(i,j) \in A}{c_{ij} x_{ij}} + \sum_{i \in N_c}\sum_{j \in N_c}c'_{ij} z_{ij}, \\
	& \mbox{subject to} \nonumber \\
	& \sum_{j \in N_c}{x_{jk}} = \sum_{j \in N_c}{x_{kj}} \qquad & \forall k \in N_d, \\
	& \sum_{j \in N_c}{x_{kj}} \leqslant \abs{P_k} \qquad & \forall k \in N_d, \\
	& \sum_{j \in N}{x_{jv}} = \sum_{j \in N}{x_{vj}} = y_v \qquad & \forall v \in N_c, \\
	& \sum_{v \in C(u)}{y_v} \geqslant 1 \qquad & \forall u \in N_c, \\
	& z_{uv} \leqslant y_v \qquad & \forall u \in N_c, \forall v \in C(u), \\
	& \sum_{j \in N}{x_{vj}} \leqslant z_{vv} \qquad & \forall v \in N_c, \\
	& \sum_{v \in C(u)}{z_{uv}} = 1 \qquad & \forall u \in N_c, \\
	& \sum_{(i,j) \in SP_{(st)}}{x_{ij}} \leqslant \abs{SP_{(st)}} - 1 \qquad & \forall s, t \in N_d, s \neq t, \forall SP_{(st)} \in SP, \\
	& \sum_{j \in N}{f_{ji}} = \sum_{j \in N_c}{d_j z_{ji}} + \sum_{j \in N}{f_{ij}} \qquad & \forall i \in N_c, \\
	& \sum_{i \in N_c}{f_{ik}} = 0 \qquad & \forall k \in N_d, \\
	& f_{ij} \leqslant (Q - d_i) x_{ij} \qquad & \forall (i, j) \in A: i \in N, j \in N_c, \\
	& d_j x_{ij} \leqslant f_{ij} \qquad & \forall (i, j) \in A: i \in N, j \in N_c, \\
	& \sum_{i \in N_c}{f_{ki}} \leqslant H \qquad & \forall k \in N_d, \\
	& x_{i,j} \in \{0,1\} \qquad & \forall (i,j) \in A, \\
	& y_v \in \{0,1\} \qquad & \forall v \in N_c, \\
	& z_{uv} \in \{0,1\} \qquad & \forall u \in N_c, \forall v \in C(u), \\
	& f_{ij} \in \mathbb{R}^+ \qquad & \forall (i,j) \in A.
\end{align}

The objective function ($16$) minimizes the total cost of the routes and allocations costs. For each depot $k \in N_d$, constraints ($17$) impose that the number of vehicles arriving $k$ must be equal to the number of vehicles leaving $k$. Constraints ($18$) bound the amount of vehicles arriving each depot. For each customer $v \in N_c$, constraints ($19$) state that the number of arcs arriving and leaving $v$ is $1$ if $v$ is visited, or $0$ otherwise. Constraints ($20$) impose that each vertex in $N_c$ must be covered by at least one route. Constraints ($21$) state that a vertex $u \in N_c$ can only be serviced through a vertex $v \in C(u)$ if $v$ is visited. Constraints ($22$) require that if a customer $v \in N_c$ is visited by a vehicle, then its demand is serviced by itself. Constraints ($23$) ensure that every vertex $u \in N_c$ is serviced by a vertex $v \in C(u)$. Constraints ($24$) prevent simple paths between depots, forcing each route to start and end in the same depot. Constraints ($25$) impose the flow conservation on each customer $i$. Constraints ($26$) ensure that the vehicle load is zero when returning to the depot. Constraints ($27$) and constraints ($28$) bound the vehicle load when traversing arc $(i, j)$. Constraints ($29$) impose that the capacity of each depot is at most $H$.

The number of decision variables used in the new MILP formulation is $O(V^2)$ and the number of decision variables used in the model by Allahyari et al. \cite{allahyari2015hybrid} is $O(V^3)$. An $O(V^2)$ algorithm can separate constraints ($24$) for integer solutions using a lazy constraint strategy. Given a graph induced by an integer solution, the separation of constraints ($24$) is performed using Depth-First-Search (DFS) \cite{cormen2022introduction}. For every pair $(i,j)$ such that $i,j \in N_d$, a DFS is performed starting from $i$. If $j$ is reached, the edges from the path between $i$ and $j$ are retrieved and then a constraint ($24$) is added to the formulation.

\section{BRKGA for Capacitated Covering Salesman Problem} \label{sec:brkga}

Guided by the Darwinian principle of the survival of the fittest, the Biased Random Key Genetic Algorithm (BRKGA) \cite{gonccalves2011biased} is an evolutionary metaheuristic in which a population of individuals, representing solutions of a combinatorial optimization problem, evolves towards the optimal.

Each individual is represented by a chromosome encoded as a vector, in which each allele is a random key uniformly drawn over the interval $[0, 1)$. The decoder method is the problem-specific component of the BRKGA which is responsible for mapping a chromosome into a solution.

An initial population of random chromosomes is created and forced into a selective pressure environment in which the best individuals are more likely to survive throughout the generations producing offsprings. 

The BRKGA partitions the population into \textit{elite} and \textit{non-elite} sets, with sizes determined by fixed parameters. The elite set is composed by the best individuals; all the remaining individuals form the non-elite set, including the \textit{mutants}, i.e., random chromosomes introduced into the population as a form of diversification. 

In each generation, the BRKGA executes the following steps:

\begin{enumerate}
    \item Decode the chromosomes, evaluating their fitness;
    \item Identify the best individuals to form the elite set;
    \item Preserve the elite set into the population for the next generation;
    \item Introduce the mutants in the next generation;
    \item Generate offsprings through the crossover of elite and non-elite chromosomes, inserting them in the next generation.
\end{enumerate}

The BRKGA has demonstrated its efficacy as a robust method for addressing various routing problems \cite{ruiz2019solving, kummer2020biased, abreu2021new, kummer2022biased, dominguez2023capacitated}. In this sense, the following sections describe how the BRKGA can be employed to solve the CCSP.

\subsection{Solution encoding}

The solution is encoded as a vector $\mathcal{X} = (x_{1}, ..., x_{n})$ of size $n = \abs{V_d}$, where $x_{i}$ is a random number in the interval $[0, 1)$, for $i = 1,\ldots, n$. Each element of $\mathcal{X}$ represents a vertex of $V_d$. 

\subsection{Decoder function}

The decoder function takes as input a vector $\mathcal{X}$ and returns a feasible solution for the CCSP represented by a set of routes $\mathcal{R}$. Let $\mathcal{X}^{'}$ be the vector resulting by sorting the keys of $\mathcal{X}$ in non-decreasing order.

The proposed decoder for the CCSP has two phases, described in the following sections.

\subsubsection{First phase - Best Fit Algorithm}

\begin{figure}[htbp]
	\centering
	\includegraphics[width=0.75\textwidth]{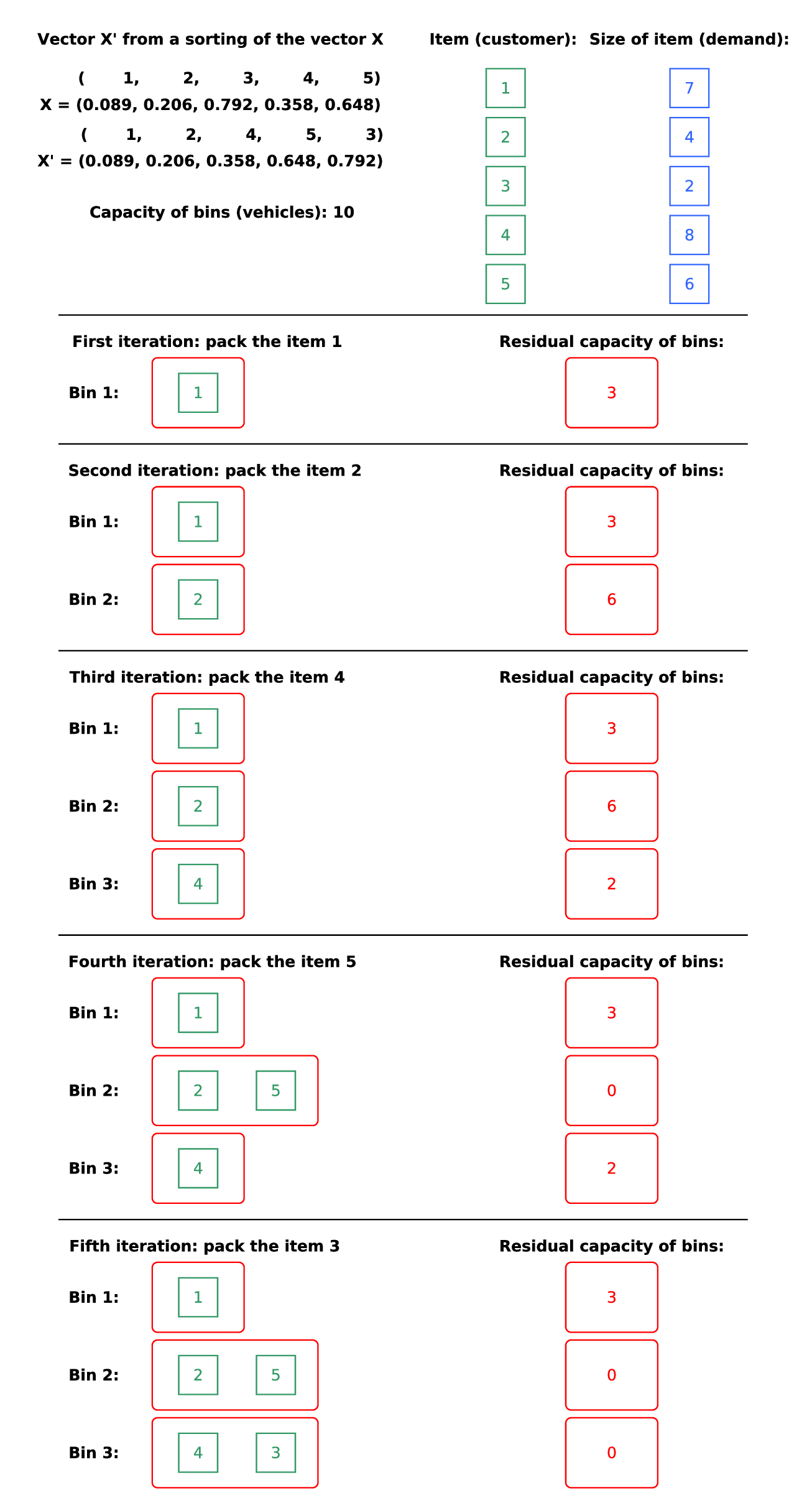} \\ [0.25cm]
	\caption{Example of Best Fit Algorithm.}
	\label{fig:exampleBestFit}
\end{figure}

The minimum number of vehicles required to service all demands can be determined by solving a \textit{Bin Packing Problem} (BPP) \cite{KellererPferschyPisinger2004}, which is NP-hard. Our decoder assigns vertices from $V_d$ to vehicles by solving the BPP approximately, using the Best Fit Algorithm (BFA). The BFA assigns each vertex to a vehicle with the least residual capacity that can still service the vertex; if no such vehicle exists, a new one is assigned (Figure~\ref{fig:exampleBestFit}).

Algorithm~\ref{alg:bestFitHeuristic} presents the BFA pseudo-code applied in the decoder, which can be implemented using self-balancing search trees leading to a worst-time complexity of $O(n \lg n)$.

\begin{algorithm}
\caption{Best fit algorithm}
\label{alg:bestFitHeuristic}
\textbf{Input:} a vector $\mathcal{X}^{'}$. \\
\textbf{Output:} a set of vehicles $\mathcal{M} = \{1,\ldots,M\}$ and their assigned vertices $\mathcal{A} = \{A_1, \ldots, A_M \}$. \\
\begin{algorithmic}[1]
{
    \State $m \leftarrow 0; \mathcal{M} \leftarrow \{ \emptyset \}$; $\mathcal{A} \leftarrow \{ \emptyset \}$;
	\For{each $x'_{i} \in \mathcal{X}^{'}$}
	    \State $v \leftarrow getVertex(i)$ -- returns the vertex associated to element $x'_{i}$;
	    \If{$\left( \exists m \in \mathcal{M} : \displaystyle \sum_{u \in A_m}{d_u} \leqslant Q - d_{v} \right)$}
	        \State $\displaystyle m \leftarrow \arg \max_{m' \in \mathcal{M}}{ \left\{ \sum_{u \in A_m'}{d_u} : \sum_{u \in A_m'}{d_u} \leqslant Q - d_{v}  \right\} }$;
	        \State $A_m \leftarrow A_m \cup \{v\}$;
	    \Else
	        \State $m \leftarrow m + 1$;
	        \State $\mathcal{M} \leftarrow \mathcal{M} \cup \{m\}$;
	        \State $A_m \leftarrow \{ v \}$;
	        \State $\mathcal{A} \leftarrow \mathcal{A} \cup \{A_m\}$
            \EndIf
        \EndFor
}
\end{algorithmic}
\end{algorithm}

\subsubsection{Second phase - route construction}

Consider that route $R_m$ of vehicle $m$ is a sequence of vertices represented as $R_m = \{R_m(0), \ldots, R_m(r_m+1) \}$, where $R_m(i)$ and $r_m$ are, respectively, the $i$-th vertex visited by vehicle $m$ and the number of vertices in $V_0$ visited by vehicle $m$. The route starts and ends at the depot, i.e., $R_m(0) = R_m(r_m+1) = v_0$.

The second phase of the decoder creates a route for each vehicle by using the following insertion cost function,
\begin{align*}
g(v,R_m) = \min_{i=\{0,\ldots,r_m\}}{\left\{ c_{(u,v)} + c_{(v,w)} - c_{(u,w)} : u = R_m(i), w = R_m(i+1) \right\}},
\end{align*}
which gives the minimum cost of inserting a vertex $v$ into a route $R_m$.

For each vehicle $m \in \mathcal{M}$ and each vertex $u \in A_m$, all unvisited vertices $v \in C(u)$ are considered to be included in route $R_m$ by checking the value of $g(v,R_m)$ and taking the vertex resulting in the least cost increment.

Algorithm~\ref{alg:routeHeuristic} presents the route construction pseudo-code used in our decoder.

\begin{algorithm}
\caption{Construction of Routes}
\label{alg:routeHeuristic}
\hspace*{\algorithmicindent} \textbf{Input:} a set of vehicles $\mathcal{M} = \{1,\ldots,M\}$ and their assigned vertices $\mathcal{A} = \{A_1, \ldots, A_M \}$. \\
\hspace*{\algorithmicindent} \textbf{Output:} a set of routes $\mathcal{R} = \{R_1, \ldots, R_m \}$. \\
\begin{algorithmic}[1]
{
    \State $\mathcal{R} \leftarrow \emptyset$; \\
    \For{each vehicle $m \in \mathcal{M}$}
        \State $R_m(0) \leftarrow \{ v_0 \}$; \\
        \State $r_m \leftarrow 0$ \\
        \For{each $v \in A_m$}
            \State ${\displaystyle S \leftarrow \left\{u \in C(v) : u \neq R_{m'}(u'), \forall m' \in \{1,\ldots,m\}, \forall u' \in \{1,\ldots,r_{m'}\}\right\}}$ -- set of unvisited candidates to service $v$ \\
	    \State $\displaystyle u \leftarrow \argmin_{u' \in S} \left\{g(u',R_m)\right\}$ -- find best candidate \\
	      \State $insert(u,R_m)$ -- insert $u$ in the best position of route $R_m$ \\
	      \State $r_m \leftarrow r_m + 1$ \\
	\EndFor
	  \State $R_m(r_m) \leftarrow \{ v_0 \}$; \\
	\State $\mathcal{R} \leftarrow \mathcal{R} \cup \{ R_m\}$; \\
    \EndFor
}
\end{algorithmic}
\end{algorithm}

A vertex $v$ is called \textit{redundant} if when removed from a route it does not change the solution feasibility. It occurs when the vertices serviced by $v$ can all be serviced by other vertices in the solution without violating the capacity of any involved vehicle. After applying Algorithm~\ref{alg:routeHeuristic}, the decoder greedily removes redundancies by considering their cost decrease, until a maximal set of redundancies is removed.

\subsection{Intra-Route}

Once the BRKGA stopping criteria is triggered, the Lin-Kernighan (LK) heuristic \cite{lin1973effective} performs the final intra-routes improvements. The LK heuristic is based on $k$-opt neighborhood, which consists in applying up to $k$ edge exchanges, and it is considered one of the best local searches for the Traveling Salesman Problem.

\subsection{Inter-Route Intensification} \label{sec:coverModel}

Following the ideas of Sartori and Buriol \cite{sartori2018matheuristic}, this paper proposes a matheuristic for the CCSP using a formulation with covering and packing constraints. Let $F$ be the set of all CCSP feasible routes, $a_{if}$ the covering matrix where for each pair $(i,f)$, $i \in V_d$ and $f \in F$, $a_{if} = 1$ if and only if vertex $i$ is serviced by route $f$, and $b_
{if}$ the visiting matrix where for each pair $(i,f)$, $i \in V_0$ and $f \in F$, $b_{if} = 1$ if and only if vertex $i$ is visited by route $f$. The formulation includes the binary variable $\lambda_f$, which denotes whether the feasible route $f \in F$ is used ($1$) or not ($0$).

The formulation reads as follows:

\begin{align} \label{eq:coverModel}
	& (Matheuristic) \nonumber \\
	& MIN \quad \sum_{f \in F}{c_f \lambda_f}, \\
	& \mbox{subject to} \nonumber \\
	& \sum_{f \in F}{a_{if} \lambda_f} \geqslant 1 \qquad & \forall i \in V_d \\
	& \sum_{f \in F}{b_{if} \lambda_f} \leqslant 1 \qquad & \forall i \in V_0 \\
	& \lambda_f \in \{0,1\} \qquad & \forall f \in F. & \nonumber
\end{align}

The objective function ($34$) minimizes the costs of the routes. Constraints ($35$) ensure that each vertex in $V_d$ must be serviced by at least one route, while constraints ($36$) impose that every vertex in $V_0$ must be visited by at most one route. Considering that the cardinality of $F$ grows exponentially, in this work we generate a pool of routes $F^{'}$ in the Matheuristic formulation. The pool of routes contains a set of CCSP feasible routes generated as follows. First, an exhaustive search is conducted to find every optimal route that services up to three vertices. All of these routes are included into $F^{'}$. The remaining routes of $F^{'}$ are filled with the elite individuals from the BRKGA generations, starting from the last generation and continuing until either the size limit of $F^{'}$ is reached or all elite individuals from each BRKGA generation are added to $F^{'}$.

\section{Computational Experiments} \label{sec:experimentsCCSP}

\subsection{Instances Benchmark}

The instances for the CCSP were obtained from the CVRP instances created by Christofides and Eilon \cite{christofides1969algorithm} and Uchoa et al. \cite{uchoa2017new}, containing between $101$ and $303$ vertices, and named as $E$-$nA$-$kB$ and $X$-$nA$-$kB$, respectively. The symbol $A$ gives the number of vertices (with the depot), and $B$ represents the number of vehicles required to service all the demands.

To generate CCSP instances, the following parameters were used:

\begin {itemize}
    \item $\abs{V_d}$: number of vertices with demand;
    \item $\abs{D(v)}$: covering size, where $v \in V_0$.
\end{itemize}

The $\abs{V_d}$ parameter varied in $10\%$, $20\%$, and $40\%$ of $n$. The vertices with demand consist of the first $\abs{V_d}$ vertices, excluding the depot, of the CVRP instance. It is worth mentioning that all demands associated with $V_d$ remain unchanged in relation to CVRP instance. Similarly, the vehicle capacity $Q$ in the CCSP instance remains the same as in the CVRP instance.

For each vertex $v \in V_0$, the set $D(v)$ is defined by the closest vertices from $v$. The cardinality of $D(v)$ varied in $7$, $9$, and $11$.

For each CVRP instance, all pairings of $\abs{V_d}$ and $\abs{D(v)}$ were considered, resulting in nine combinations and a benchmark of $495$ instances. A pre-processing was conducted in the set of instances to remove any vertex $v \in V_0$ such that $D(v) \cap V_d = \emptyset$.

The MDCTVRP instances were created by Allahyari et al. \cite{allahyari2015hybrid}, and they are divided in small ($120$ instances) and large ($160$ instances). The small instances contain up to $30$ vertices, and the large instances have up to $90$ vertices. In small instances, the vehicle capacity fluctuates between $140$ and $150$, while in large instances, the vehicle capacity alternates between $160$ and $170$. The small instances are divided into three categories, and the large instances are divided into four categories. Each category has eight different groups of instances. The instances are named InputXYZT, where ``X'', ``Y'', ``Z'', and ``T'' give the category, number of depots, vehicle capacity, and the coverage coefficient (which defines the cost of serving a vertex), respectively. Five instances were generated for each group of instances.

\subsection{Computational Settings}

The ILP and MILP formulations were implemented and solved using Gurobi 8.1.1 version. The execution time limit were set to a one hour, except for the $MDCTVRP_m$ formulation, which was set to two hours following Allahyari et al. \cite{allahyari2015hybrid}. The experiments were conducted on a PC under Ubuntu $10.12$, and CPU Intel Xeon(R) Silver $3114$ $2.20$ GHz, with $32$GB of RAM. The BRKGA developed for the CCSP used the C++ framework from Resende and Toso \cite{toso2015c++}. The parameters used by the BRKGA are listed in Table~\ref{tab:brkgaParameters}. The implementation of the LK heuristic proposed by Helsgaun \cite{helsgaun2000effective} was employed. The matheuristic adopted a size limit for the pool $F^{'}$ of $1$ million routes.

\begin{table}[htbp]
\centering
\caption{BRKGA parameters.}
\label{tab:brkgaParameters}
\begin{tabular}{cc}
\hline
Parameter                                        & Value \\ \hline
Population size                                  & 1000  \\
Fraction of population to be elite individuals   & 40\%  \\
Fraction of population to be replaced by mutants & 20\%  \\
Crossover probability                            & 70\%  \\ \hline
\end{tabular}
\end{table}

\subsection{Evaluated Methodologies}

Five methodologies were implemented and evaluated in the computational experiments:

\begin{itemize}
    \item $\mathbb{CCSP}_{1}$: solution of the $CCSP_1$ model, initially ignoring Constraints $(7)$, and later including them in the formulation using a \textit{\bfseries lazy constraint} strategy;
    \item $\mathbb{BRKGA}$: implementation of the BRKGA for the CCSP described in Section~\ref{sec:brkga};
    \item $\mathbb{CCSP}_{1s}$: same as $\mathbb{CCSP}_{1}$, however the solution obtained by the BRKGA is given as warm start for the $CCSP_1$ model;
    \item $Matheuristic$: solution of the matheuristic described in Subsection~\ref{sec:coverModel};
    \item $\mathbb{MDCTVRP}_{m}$: solution of the $MDCTVRP_m$ formulation, initially ignoring Constraints ($24$), and including them on-demand in the formulation using a \textit{\bfseries lazy constraint} strategy.
\end{itemize}

The $\mathbb{MDCTVRP}_{m}$ was compared with the flow-based formulation ($F_{flow}$) and node-based formulation ($F_{node}$), both proposed by Allahyari et al. \cite{allahyari2015hybrid}, which are the state-of-the-art exact methodologies for the MDCTVRP, to the best of our knowledge.

\subsection{Results for the CCSP}

Full experimental data, results, instances, and source codes are available on-line\footnote{\label{note2}\url{http://www.ic.unicamp.br/~fusberti/problems/ccsp}}. The results of the computational experiments show that the $\mathbb{CCSP}_{1}$ methodology was able to obtain upper bounds for $430$ out of $495$ instances. In addition, the $\mathbb{CCSP}_{1}$ methodology proved optimality for $71$ instances size up to $101$ vertices. Analyzing the results of the $\mathbb{BRKGA}$ methodology, we can note that for $407$ instances, the obtained solutions were better than the upper bounds obtained by $\mathbb{CCSP}_{1}$.

With respect to methodologies $\mathbb{CCSP}_{1s}$ and $Matheuristic$, the results show that the matheuristic was more effective to improve the solutions obtained from $\mathbb{BRKGA}$. $Matheuristic$ methodology improved the BRKGA solutions for $187$ out of $495$ instances, while $\mathbb{CCSP}_{1s}$ improved for $88$ instances. The average cost of improvements made by $\mathbb{CCSP}_{1s}$ and $Matheuristic$ on BRKGA solutions were approximately $1.69\%$ and $2.3\%$, respectively.

\begin{figure}[htbp]
	\centering
	\includegraphics[width=0.84\textwidth]{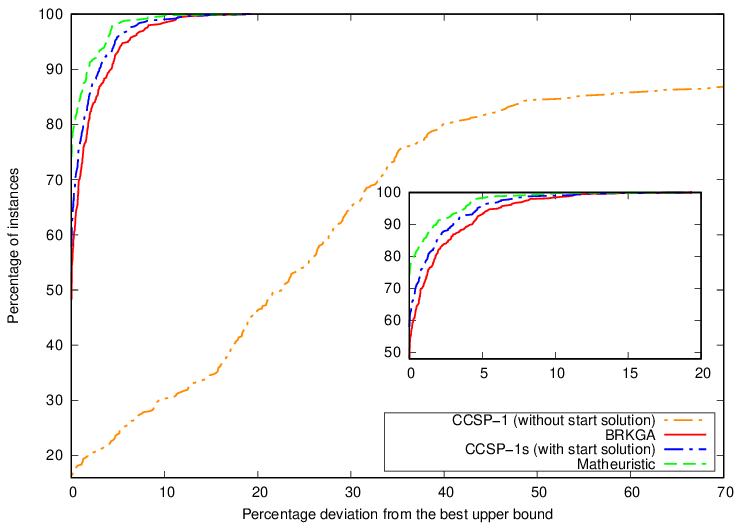} \\ [0.25cm]
	\caption{Performance profiles in terms of deviation ($\%$) from the best upper bound.}
	\label{fig:graficos_comparativos_instancias_upper_bounds}
\end{figure}

Figure~\ref{fig:graficos_comparativos_instancias_upper_bounds} shows, for each methodology, the percentage of solved instances in function of the deviation from the best upper bound ($(\frac{UB - BestUB}{UB}) * 100$). The performance profile clearly shows the $\mathbb{BRKGA}$ dominating $\mathbb{CCSP}_{1}$ with respect to upper bounds. The $\mathbb{BRKGA}$ obtained the best solutions for approximately $48\%$ of instances, while the $\mathbb{CCSP}_{1}$ obtained for approximately $16\%$ of instances. The $\mathbb{CCSP}_{1s}$ was able to improve the warm start $\mathbb{BRKGA}$ solution for several instances, obtaining the best solutions for approximately $58\%$ of instances. Finally, comparing methodologies $Matheuristic$ and $\mathbb{CCSP}_{1s}$, $Matheuristic$ was more effective in improving the BRKGA solutions. The $Matheuristic$ methodology obtained the best solution for approximately $74\%$ of instances, outperforming $\mathbb{CCSP}_{1s}$.

\subsection{Results for the MDCTVRP}

Tables~\ref{tab:mdctvrpSmallInstances} and~\ref{tab:mdctvrpLargeInstances} report the results for the small and large instances, respectively. Table~\ref{tab:mdctvrpSmallInstances} reports for each methodology and for each group of small-size instances, the average gap (Avg.gap), the number of optimal solutions (\#Opt), and the average running time (Avg.time).

For small instances, the overall optimality gaps were $0.10\%$, $8.99\%$, and $30.28\%$ for the $\mathbb{MDCTVRP}_{m}$, $F_{flow}$, and $F_{node}$ methodologies, respectively. From $120$ small instances, $117$, $10$, and $4$ optimal solutions were obtained by $\mathbb{MDCTVRP}_{m}$, $F_{flow}$, and $F_{node}$, respectively.

Table~\ref{tab:mdctvrpLargeInstances} gives the results from $\mathbb{MDCTVRP}_{m}$ for each group of large-size instances. The column group $\mathbb{MDCTVRP}_{m}$ reports the averages upper bound (Avg.ub), lower bound (Avg.lb), optimality gap (Avg.gap), running time (Avg.time), and number of optimal solutions (\#Opt). The column group GRASP x ILS \cite{allahyari2015hybrid} reports the results obtained by the hybrid meta-heuristic proposed by Allahyari et al. \cite{allahyari2015hybrid}. ``Avg'' gives the average cost obtained over five executions for each instance. Avg.gap$_{LP}$ and Avg.gap $_{\mathbb{MDCTVRP}_{m}}$ give the ``Avg'' gap from the linear relaxation of $F_{flow}$ formulation, and the average lower bound (Avg.lb) of $\mathbb{MDCTVRP}_{m}$, respectively.

It should be noticed that, due to the large number of variables, Allahyari et al. \cite{allahyari2015hybrid} have not executed experiments on large instances with $F_{flow}$. The computational results have shown that $\mathbb{MDCTVRP}_{m}$ achieved good lower bounds for large. Previously, the GRASP x ILS overall gap with respect to the linear programming relaxation of $F_{flow}$ was $20.98\%$, while with the new lower bounds obtained by $\mathbb{MDCTVRP}_{m}$ the GRASP x ILS overall gap was improved to $7.98\%$. It is worth noting that even though $\mathbb{MDCTVRP}_{m}$ is an exact methodology, the upper bounds were close to the solutions cost obtained by the GRASP x ILS. More specifically, the solutions cost obtained by GRASP x ILS are, on average, only $1.37\%$ apart from the upper bounds obtained by $\mathbb{MDCTVRP}_{m}$. 

\begin{landscape}
\makeatletter
\let\ps@plain\ps@mystyle
\makeatother
\pagestyle{fancy}
\fancyhf{}
\renewcommand{\headrulewidth}{0pt}
\setlength{\headheight}{14.5pt}
\fancyhead[L]{\thepage}
\fancypagestyle{plain}{\fancyhf{}\fancyhead[L]{\thepage}}
\begin{table}[htbp]
\scriptsize
\centering
\caption{Results of computational experiments for the small-size instances.}
\begin{tabular}{cccccccccccccc}
\label{tab:mdctvrpSmallInstances} \\
\cline{4-14}
                   &           &  & \multicolumn{3}{c}{$\mathbb{MDCTVRP}_{m}$} &  & \multicolumn{3}{c}{$F_{flow}$} &  & \multicolumn{3}{c}{$F_{node}$} \\ \cline{4-6} \cline{8-10} \cline{12-14} 
Category           & Group     &  & Avg.gap             & \#Opt            & Avg.time            &  & Avg.gap   & \#Opt   & Avg.time  &  & Avg.gap   & \#Opt   & Avg.time  \\ \hline
\multirow{8}{*}{1} & Input1000 &  & 0.00                & 5                & 8.00                &  & 6.37      & 0       & 7200.00  &  & 22.25     & 0       & 7200.00  \\
                   & Input1001 &  & 0.00                & 5                & 8.80                &  & 6.38      & 0       & 7200.00  &  & 21.29     & 0       & 7200.00  \\
                   & Input1010 &  & 0.00                & 5                & 6.60                &  & 1.74      & 2       & 6015.00  &  & 17.71     & 0       & 7200.00  \\
                   & Input1011 &  & 0.00                & 5                & 5.20                &  & 2.48      & 1       & 7066.00  &  & 18.63     & 0       & 7200.00  \\
                   & Input1100 &  & 0.00                & 5                & 7.00                &  & 4.96      & 2       & 5621.00  &  & 14.19     & 1       & 7200.00  \\
                   & Input1101 &  & 0.00                & 5                & 6.80                &  & 4.87      & 1       & 5848.00  &  & 12.44     & 1       & 7200.00  \\
                   & Input1110 &  & 0.00                & 5                & 5.80                &  & 3.14      & 2       & 5504.00  &  & 13.85     & 1       & 7200.00  \\
                   & Input1111 &  & 0.00                & 5                & 6.20                &  & 2.03      & 2       & 5220.00  &  & 11.57     & 1       & 7200.00  \\
                   &           &  &                     &                  &                     &  &           &         &           &  &           &         &           \\
\multirow{8}{*}{2} & Input2000 &  & 0.00                & 5                & 140.20              &  & 10.56     & 0       & 7200.00  &  & 42.30     & 0       & 7200.00  \\
                   & Input2001 &  & 0.00                & 5                & 184.60              &  & 10.81     & 0       & 7200.00  &  & 38.64     & 0       & 7200.00  \\
                   & Input2010 &  & 0.00                & 5                & 70.20               &  & 10.39     & 0       & 7200.00  &  & 37.45     & 0       & 7200.00  \\
                   & Input2011 &  & 0.00                & 5                & 115.40              &  & 11.14     & 0       & 7200.00  &  & 33.94     & 0       & 7200.00  \\
                   & Input2100 &  & 0.00                & 5                & 66.60               &  & 7.98      & 0       & 7200.00  &  & 31.33     & 0       & 7200.00  \\
                   & Input2101 &  & 0.00                & 5                & 138.20              &  & 7.63      & 0       & 7200.00  &  & 27.85     & 0       & 7200.00  \\
                   & input2110 &  & 0.00                & 5                & 73.20               &  & 9.02      & 0       & 7200.00  &  & 28.45     & 0       & 7200.00  \\
                   & Input2111 &  & 0.00                & 5                & 44.60               &  & 9.55      & 0       & 7200.00  &  & 27.43     & 0       & 7200.00  \\
                   &           &  &                     &                  &                     &  &           &         &           &  &           &         &           \\
\multirow{8}{*}{3} & Input3000 &  & 1.01                & 4                & 2653.20            &  & 14.48     & 0       & 7200.00  &  & 50.29     & 0       & 7200.00  \\
                   & Input3001 &  & 0.71                & 4                & 2853.20            &  & 14.93     & 0       & 7200.00  &  & 44.15     & 0       & 7200.00  \\
                   & Input3010 &  & 0.00                & 5                & 1865.80            &  & 15.23     & 0       & 7200.00  &  & 44.46     & 0       & 7200.00  \\
                   & Input3011 &  & 0.58                & 4                & 1923.60            &  & 14.29     & 0       & 7200.00  &  & 41.20     & 0       & 7200.00  \\
                   & Input3100 &  & 0.00                & 5                & 288.20              &  & 12.95     & 0       & 7200.00  &  & 39.75     & 0       & 7200.00  \\
                   & Input3101 &  & 0.00                & 5                & 246.40              &  & 11.83     & 0       & 7200.00  &  & 35.90     & 0       & 7200.00  \\
                   & Input3110 &  & 0.00                & 5                & 383.60              &  & 12.25     & 0       & 7200.00  &  & 36.70     & 0       & 7200.00  \\
                   & Input3111 &  & 0.00                & 5                & 492.20              &  & 10.85     & 0       & 7200.00  &  & 35.00     & 0       & 7200.00  \\ \hline
Average            &           &  & 0.10                &                  & 483.07              &  & 8.99      &         & 6869.75  &  & 30.28     &         & 7200.00  \\ \hline
\end{tabular}
\end{table}
\end{landscape}

\begin{landscape}
\makeatletter
\let\ps@plain\ps@mystyle
\makeatother
\pagestyle{fancy}
\fancyhf{}
\renewcommand{\headrulewidth}{0pt}
\setlength{\headheight}{14.5pt}
\fancyhead[L]{\thepage}
\fancypagestyle{plain}{\fancyhf{}\fancyhead[L]{\thepage}}
\begin{table}[htbp]
\renewcommand{\arraystretch}{0.92}
\scriptsize
\centering
\caption{Results of computational experiments for the large-size instances.}
\begin{tabular}{cccccccccccc}
\label{tab:mdctvrpLargeInstances} \\
\cline{4-12}
                   &           &  & \multicolumn{5}{c}{$\mathbb{MDCTVRP}_{m}$} &  & \multicolumn{3}{c}{GRASP x ILS \cite{allahyari2015hybrid}}                                                                                             \\ \cline{4-8} \cline{10-12} 
Category           & Group     &  & Avg.ub      & Avg.lb    & Avg.gap    & \#Opt    & Avg.time   &  & Avg     & Avg.gap$_{LP}$ & Avg.gap$_{\mathbb{MDCTVRP}_{m}}$ \\ \hline
\multirow{8}{*}{4} & Input4000 &  & 799.45      & 751.98    & 6.22       & 0        & 7200.00   &  & 796.21  & 19.30                                          & 5.88                                                             \\
                   & Input4001 &  & 808.23      & 764.91    & 5.55       & 0        & 7200.00   &  & 806.95  & 18.57                                          & 5.50                                                             \\
                   & Input4010 &  & 787.18      & 729.22    & 7.86       & 0        & 7200.00   &  & 775.66  & 19.68                                          & 6.37                                                             \\
                   & Input4011 &  & 792.13      & 744.53    & 6.26       & 0        & 7200.00   &  & 787.64  & 19.12                                          & 5.79                                                             \\
                   & Input4100 &  & 737.27      & 692.92    & 6.24       & 0        & 7200.00   &  & 733.72  & 21.90                                          & 5.89                                                             \\
                   & Input4101 &  & 750.94      & 701.21    & 6.98       & 0        & 7200.00   &  & 745.62  & 20.91                                          & 6.33                                                             \\
                   & Input4110 &  & 721.12      & 678.30    & 6.09       & 1        & 6650.80   &  & 711.94  & 21.26                                          & 4.96                                                             \\
                   & Input4111 &  & 730.01      & 689.79    & 5.72       & 0        & 7200.00   &  & 728.17  & 21.03                                          & 5.56                                                             \\
                   &           &  &             &           &            &          &            &  &         &                                                &                                                                  \\
\multirow{8}{*}{5} & Input5000 &  & 899.58      & 815.45    & 10.27      & 0        & 7200.00   &  & 880.01  & 19.80                                          & 7.92                                                             \\
                   & Input5001 &  & 903.58      & 829.00    & 8.95       & 0        & 7200.00   &  & 895.99  & 18.93                                          & 8.08                                                             \\
                   & Input5010 &  & 869.08      & 784.04    & 10.82      & 0        & 7200.00   &  & 857.27  & 21.41                                          & 9.34                                                             \\
                   & Input5011 &  & 886.07      & 800.62    & 10.63      & 0        & 7200.00   &  & 873.99  & 20.46                                          & 9.16                                                             \\
                   & Input5100 &  & 810.75      & 732.59    & 10.65      & 0        & 7200.00   &  & 796.88  & 22.12                                          & 8.78                                                             \\
                   & Input5101 &  & 827.35      & 748.43    & 10.52      & 0        & 7200.00   &  & 813.70  & 20.85                                          & 8.72                                                             \\
                   & Input5110 &  & 785.86      & 711.22    & 10.51      & 0        & 7200.00   &  & 777.99  & 23.23                                          & 9.39                                                             \\
                   & Input5111 &  & 804.80      & 728.69    & 10.37      & 0        & 7200.00   &  & 794.78  & 21.71                                          & 9.07                                                             \\
                   &           &  &             &           &            &          &            &  &         &                                                &                                                                  \\
\multirow{8}{*}{6} & Input6000 &  & 1005.58    & 922.85    & 8.88       & 0        & 7200.00   &  & 997.18  & 20.01                                          & 8.05                                                             \\
                   & Input6001 &  & 1037.50    & 936.43    & 10.75      & 0        & 7200.00   &  & 1016.10 & 19.43                                          & 8.51                                                             \\
                   & Input6010 &  & 987.04      & 898.72    & 9.63       & 0        & 7200.00   &  & 968.81  & 20.59                                          & 7.80                                                             \\
                   & Input6011 &  & 998.51      & 916.29    & 8.83       & 0        & 7200.00   &  & 989.60  & 20.16                                          & 8.00                                                             \\
                   & Input6100 &  & 955.65      & 875.82    & 9.07       & 0        & 7200.00   &  & 946.88  & 20.63                                          & 8.11                                                             \\
                   & Input6101 &  & 984.79      & 893.52    & 10.09      & 0        & 7200.00   &  & 967.23  & 20.11                                          & 8.25                                                             \\
                   & Input6110 &  & 936.80      & 848.97    & 10.27      & 0        & 7200.00   &  & 920.24  & 21.58                                          & 8.40                                                             \\
                   & Input6111 &  & 953.62      & 867.56    & 9.88       & 0        & 7200.00   &  & 939.51  & 20.77                                          & 8.29                                                             \\
                   &           &  &             &           &            &          &            &  &         &                                                &                                                                  \\
\multirow{8}{*}{7} & Input7000 &  & 1031.77    & 930.53    & 10.79      & 0        & 7200.00   &  & 1013.92 & 21.35                                          & 8.96                                                             \\
                   & Input7001 &  & 1047.06    & 935.57    & 11.88      & 0        & 7200.00   &  & 1019.34 & 20.89                                          & 8.95                                                             \\
                   & Input7010 &  & 1012.71    & 905.58    & 11.78      & 0        & 7200.00   &  & 989.22  & 21.90                                          & 9.24                                                             \\
                   & Input7011 &  & 1018.47    & 910.80    & 11.83      & 0        & 7200.00   &  & 995.08  & 21.47                                          & 9.25                                                             \\
                   & Input7100 &  & 933.58      & 842.19    & 10.81      & 0        & 7200.00   &  & 918.91  & 22.90                                          & 9.11                                                             \\
                   & Input7101 &  & 948.35      & 849.03    & 11.65      & 0        & 7200.00   &  & 924.61  & 22.39                                          & 8.90                                                             \\
                   & Input7110 &  & 917.68      & 830.41    & 10.51      & 0        & 7200.00   &  & 904.92  & 23.56                                          & 8.97                                                             \\
                   & Input7111 &  & 919.48      & 832.89    & 10.37      & 0        & 7200.00   &  & 915.05  & 23.24                                          & 9.86                                                             \\ \hline
Average            &           &  & 893.81      &     815.63      & 9.40       &          &            &  & 881.35  & 20.98                                          & 7.98                                                             \\ \hline
\end{tabular}
\end{table}
\end{landscape}

\section{Final Remarks} \label{sec:conclusionsCCSP}

This work proposes the Capacitated Covering Salesman Problem (CCSP), a problem that approaches the notion of coverage in vehicle routing problems. Two ILP formulations and a BRKGA are proposed to solve the CCSP. From the set of instances for the CVRP \cite{christofides1969algorithm,uchoa2017new}, a benchmark of instances for CCSP was generated.

Computational experiments conducted on a benchmark of $198$ instances for CCSP evaluated the ILP formulation and the BRKGA. The results show the effectiveness of the BRKGA in obtaining upper bounds for all instances. The $CCSP_1$ obtained optimal solutions for $71$ instances, with up to $101$ vertices.

Furthermore, a new MILP formulation is proposed for the Multi-Depot Covering Tour Vehicle Routing Problem (MDCTVRP). Computational experiments were conducted on a benchmark of $280$ instances from literature. The overall results show unequivocally the new formulation outperforming the best known exact methodology from literature, obtaining $118$ new optimal solutions and improving all known lower bounds.

Future works should focus on valid inequalities and a branch-and-cut framework for the solution of CCSP and MDCTVRP. Another promising field of research is to consider multi-objective vehicle routing problem with covering range being the additional objective function.

\section*{Acknowledgments}

This work was supported by CAPES, CNPq, and Fapesp (grants 140960/2017-1, 314384/2018-9, 435520/2018-0, 2015/11937-9).

\bibliographystyle{unsrt}  
\bibliography{references}  

\begin{thebibliography}{10}

\bibitem{dantzig1959truck}
George~B Dantzig and John~H Ramser.
\newblock The truck dispatching problem.
\newblock {\em Management science}, 6(1):80--91, 1959.

\bibitem{cordeau2007vehicle}
Jean-Fran{\c{c}}ois Cordeau, Gilbert Laporte, Martin~WP Savelsbergh, and Daniele Vigo.
\newblock Vehicle routing.
\newblock {\em Handbooks in operations research and management science}, 14:367--428, 2007.

\bibitem{golden2008vehicle}
Bruce~L Golden, Subramanian Raghavan, and Edward~A Wasil.
\newblock {\em The vehicle routing problem: latest advances and new challenges}, volume~43.
\newblock Springer Science \& Business Media, 2008.

\bibitem{allahyari2015hybrid}
Somayeh Allahyari, Majid Salari, and Daniele Vigo.
\newblock A hybrid metaheuristic algorithm for the multi-depot covering tour vehicle routing problem.
\newblock {\em European Journal of Operational Research}, 242(3):756--768, 2015.

\bibitem{current1989coveringa2}
John~R Current and David~A Schilling.
\newblock The covering salesman problem.
\newblock {\em Transportation science}, 23(3):208--213, 1989.

\bibitem{applegate2007a2}
David~L. Applegate, Robert~E. Bixby, Vasek Chvatal, and William~J. Cook.
\newblock {\em The Traveling Salesman Problem: A Computational Study (Princeton Series in Applied Mathematics)}.
\newblock Princeton University Press, Princeton, NJ, USA, 2007.

\bibitem{golden2012generalizeda2}
Bruce Golden, Zahra Naji-Azimi, S~Raghavan, Majid Salari, and Paolo Toth.
\newblock The generalized covering salesman problem.
\newblock {\em INFORMS Journal on Computing}, 24(4):534--553, 2012.

\bibitem{gendreau1997coveringa2}
Michel Gendreau, Gilbert Laporte, and Fr{\'e}d{\'e}ric Semet.
\newblock The covering tour problem.
\newblock {\em Operations Research}, 45(4):568--576, 1997.

\bibitem{hachicha2000heuristics}
Mondher Hachicha, M~John Hodgson, Gilbert Laporte, and Fr{\'e}d{\'e}ric Semet.
\newblock Heuristics for the multi-vehicle covering tour problem.
\newblock {\em Computers \& Operations Research}, 27(1):29--42, 2000.

\bibitem{naji2012covering}
Zara Naji-Azimi, Jacques Renaud, Angel Ruiz, and Majid Salari.
\newblock A covering tour approach to the location of satellite distribution centers to supply humanitarian aid.
\newblock {\em European Journal of Operational Research}, 222(3):596--605, 2012.

\bibitem{ha2013exact}
Minh~Hoang Ha, Nathalie Bostel, Andr{\'e} Langevin, and Louis-Martin Rousseau.
\newblock An exact algorithm and a metaheuristic for the multi-vehicle covering tour problem with a constraint on the number of vertices.
\newblock {\em European Journal of Operational Research}, 226(2):211--220, 2013.

\bibitem{murakami2014column}
Keisuke Murakami.
\newblock A column generation approach for the multi-vehicle covering tour problem.
\newblock In {\em Automation Science and Engineering (CASE), 2014 IEEE International Conference on}, pages 1063--1068. IEEE, 2014.

\bibitem{jozefowiez2014branch}
Nicolas Jozefowiez.
\newblock A branch-and-price algorithm for the multivehicle covering tour problem.
\newblock {\em Networks}, 64(3):160--168, 2014.

\bibitem{kammoun2017integration}
Manel Kammoun, Houda Derbel, Mostapha Ratli, and Bassem Jarboui.
\newblock An integration of mixed vnd and vns: the case of the multivehicle covering tour problem.
\newblock {\em International Transactions in Operational Research}, 24(3):663--679, 2017.

\bibitem{tillman1969multiple}
Frank~A Tillman.
\newblock The multiple terminal delivery problem with probabilistic demands.
\newblock {\em Transportation Science}, 3(3):192--204, 1969.

\bibitem{cormen2022introduction}
Thomas~H Cormen, Charles~E Leiserson, Ronald~L Rivest, and Clifford Stein.
\newblock {\em Introduction to algorithms}.
\newblock MIT press, 2022.

\bibitem{gonccalves2011biased}
Jos{\'e}~Fernando Gon{\c{c}}alves and Mauricio~GC Resende.
\newblock Biased random-key genetic algorithms for combinatorial optimization.
\newblock {\em Journal of Heuristics}, 17(5):487--525, 2011.

\bibitem{ruiz2019solving}
Efrain Ruiz, Valeria Soto-Mendoza, Alvaro Ernesto~Ruiz Barbosa, and Ricardo Reyes.
\newblock Solving the open vehicle routing problem with capacity and distance constraints with a biased random key genetic algorithm.
\newblock {\em Computers \& Industrial Engineering}, 133:207--219, 2019.

\bibitem{kummer2020biased}
Alberto~F Kummer~N, Luciana~S Buriol, and Olinto~CB de~Ara{\'u}jo.
\newblock A biased random key genetic algorithm applied to the vrptw with skill requirements and synchronization constraints.
\newblock In {\em Proceedings of the 2020 Genetic and Evolutionary Computation Conference}, pages 717--724, 2020.

\bibitem{abreu2021new}
Levi~R Abreu, Roberto~F Tavares-Neto, and Marcelo~S Nagano.
\newblock A new efficient biased random key genetic algorithm for open shop scheduling with routing by capacitated single vehicle and makespan minimization.
\newblock {\em Engineering Applications of Artificial Intelligence}, 104:104373, 2021.

\bibitem{kummer2022biased}
Alberto~F Kummer, Olinto~CB de~Ara{\'u}jo, Luciana~S Buriol, and Mauricio~GC Resende.
\newblock A biased random-key genetic algorithm for the home health care problem.
\newblock {\em International Transactions in Operational Research}, 2022.

\bibitem{dominguez2023capacitated}
Sa{\'u}l Dom{\'\i}nguez-Casasola, Jos{\'e}~Luis Gonz{\'a}lez-Velarde, Yasm{\'\i}n~{\'A} R{\'\i}os-Sol{\'\i}s, and Kevin~Alain Reyes-Vega.
\newblock The capacitated family traveling salesperson problem.
\newblock {\em International Transactions in Operational Research}, 2023.

\bibitem{KellererPferschyPisinger2004}
H~Kellerer, U~Pferschy, and D~Pisinger.
\newblock {\em Knapsack Problems}, volume~1.
\newblock Springer-Verlag Berlin Heidelberg, 2004.

\bibitem{lin1973effective}
Shen Lin and Brian~W Kernighan.
\newblock An effective heuristic algorithm for the traveling-salesman problem.
\newblock {\em Operations research}, 21(2):498--516, 1973.

\bibitem{sartori2018matheuristic}
Carlo~S Sartori and Luciana~S Buriol.
\newblock A matheuristic approach to the pickup and delivery problem with time windows.
\newblock In {\em International Conference on Computational Logistics}, pages 253--267. Springer, 2018.

\bibitem{christofides1969algorithm}
Nicos Christofides and Samuel Eilon.
\newblock An algorithm for the vehicle-dispatching problem.
\newblock {\em Journal of the Operational Research Society}, 20(3):309--318, 1969.

\bibitem{uchoa2017new}
Eduardo Uchoa, Diego Pecin, Artur Pessoa, Marcus Poggi, Thibaut Vidal, and Anand Subramanian.
\newblock New benchmark instances for the capacitated vehicle routing problem.
\newblock {\em European Journal of Operational Research}, 257(3):845--858, 2017.

\bibitem{toso2015c++}
Rodrigo~F Toso and Mauricio~GC Resende.
\newblock A c++ application programming interface for biased random-key genetic algorithms.
\newblock {\em Optimization Methods and Software}, 30(1):81--93, 2015.

\bibitem{helsgaun2000effective}
Keld Helsgaun.
\newblock An effective implementation of the lin--kernighan traveling salesman heuristic.
\newblock {\em European Journal of Operational Research}, 126(1):106--130, 2000.

\end{thebibliography}

\end{document}